%% file: acl2023.tex
\pdfoutput=1

\documentclass[11pt]{article}

\usepackage[]{ACL2023}

\usepackage{times}
\usepackage{latexsym}

\usepackage[T1]{fontenc}

\usepackage[utf8]{inputenc}

\usepackage{microtype}

\usepackage{inconsolata}

\usepackage{multirow}
\usepackage{graphicx} 
\usepackage{booktabs}
\usepackage{xcolor,colortbl}
\usepackage{color}
\usepackage[most]{tcolorbox}
\usepackage{pifont}
\usepackage[figuresright]{rotating}
\usepackage{mathrsfs}
\usepackage{amsmath}
\usepackage{pifont}
\usepackage[figuresright]{rotating}
\usepackage{algorithm}
\usepackage{algpseudocode}
\definecolor{MyGreen}{RGB}{70,186,110}

\makeatletter
\def\therule{\makebox[\algorithmicindent][l]{\hspace*{.5em}\vrule height .75\baselineskip depth .25\baselineskip}}%

\newtoks\therules
\therules={}
\def\appendto#1#2{\expandafter#1\expandafter{\the#1#2}}
\def\gobblefirst#1{
  #1\expandafter\expandafter\expandafter{\expandafter\@gobble\the#1}}%
\def\LState{\State\unskip\the\therules}
\def\pushindent{\appendto\therules\therule}%
\def\popindent{\gobblefirst\therules}%
\def\printindent{\unskip\the\therules}%
\def\printandpush{\printindent\pushindent}%
\def\popandprint{\popindent\printindent}%

\algdef{SE}[WHILE]{While}{EndWhile}[1]
  {\printandpush\algorithmicwhile\ #1\ \algorithmicdo}
  {\popandprint\algorithmicend\ \algorithmicwhile}%
\algdef{SE}[FOR]{For}{EndFor}[1]
  {\printandpush\algorithmicfor\ #1\ \algorithmicdo}
  {\popandprint\algorithmicend\ \algorithmicfor}%
\algdef{S}[FOR]{ForAll}[1]
  {\printindent\algorithmicforall\ #1\ \algorithmicdo}%
\algdef{SE}[LOOP]{Loop}{EndLoop}
  {\printandpush\algorithmicloop}
  {\popandprint\algorithmicend\ \algorithmicloop}%
\algdef{SE}[REPEAT]{Repeat}{Until}
  {\printandpush\algorithmicrepeat}[1]
  {\popandprint\algorithmicuntil\ #1}%
\algdef{SE}[IF]{If}{EndIf}[1]
  {\printandpush\algorithmicif\ #1\ \algorithmicthen}
  {\popandprint\algorithmicend\ \algorithmicif}%
\algdef{C}[IF]{IF}{ElsIf}[1]
  {\popandprint\pushindent\algorithmicelse\ \algorithmicif\ #1\ \algorithmicthen}%
\algdef{Ce}[ELSE]{IF}{Else}{EndIf}
  {\popandprint\pushindent\algorithmicelse}%
\algdef{SE}[PROCEDURE]{Procedure}{EndProcedure}[2]
   {\printandpush\algorithmicprocedure\ \textproc{#1}\ifthenelse{\equal{#2}{}}{}{(#2)}}%
   {\popandprint\algorithmicend\ \algorithmicprocedure}%
\algdef{SE}[FUNCTION]{Function}{EndFunction}[2]
   {\printandpush\algorithmicfunction\ \textproc{#1}\ifthenelse{\equal{#2}{}}{}{(#2)}}%
   {\popandprint\algorithmicend\ \algorithmicfunction}%
\makeatother

\newtcolorbox[list inside=prompt,auto counter,number within=section]{prompt}[1][]{
    colbacktitle=black!60,
    coltitle=white,
    fontupper=\footnotesize,
    boxsep=5pt,
    left=0pt,
    right=0pt,
    top=0pt,
    bottom=0pt,
    boxrule=1pt,
    #1,
}

\newcommand{\firstc}{{\cellcolor[RGB]{ 171, 235, 198 }}}
\newcommand{\secondc}{{\cellcolor[RGB]{ 253, 235, 208}}}

\title{\textit{Learning to Break}:\\ Knowledge-Enhanced Reasoning in Multi-Agent Debate System}

\author{Haotian Wang$^1$, Xiyuan Du$^1$, Weijiang Yu$^2$\thanks{~~Corresponding author.} , Qianglong Chen$^3$\\ 
\textbf{Kun Zhu$^1$, Zheng Chu$^1$, Lian Yan$^1$, Yi Guan$^1$\footnotemark[1]}\\
  $^1$Harbin Institute of Technology \quad $^2$Sun Yat-sen University \quad
  $^3$ Zhejiang University \\
  \texttt{\{wanght1998,weijiangyu8,chenqianglong.ai\}@gmail.com},
  \texttt{guanyi@hit.edu.cn} \\
  \texttt{\{xydu,kzhu,zchu\}@ir.hit.edu.cn}, \texttt{yanlian0216@163.com}
}

\begin{document}
\maketitle

\input{sections/abstract}
\input{sections/1_introduction}
\input{sections/2_relatedwork}
\input{sections/3_method}

\input{sections/4_experiments}

\input{sections/5_results}
\input{sections/6_conclusion}

\bibliography{custom}
\bibliographystyle{acl_natbib}

\input{sections/appendix}

\end{document}

%% file: sections/abstract.tex
\begin{abstract}
    Multi-agent debate system (MAD) imitating the process of human discussion in pursuit of truth, aims to align the correct cognition of different agents for the optimal solution. It is challenging to make various agents perform right and highly consistent cognition due to their limited and different knowledge backgrounds (i.e., cognitive islands), which hinders the search for the optimal solution. 
    To address the challenge, we propose a novel \underline{M}ulti-\underline{A}gent \underline{D}ebate with \underline{K}nowledge-\underline{E}nhanced framework (\textbf{MADKE}) to promote the system to find the solution. First, we involve a shared retrieval knowledge pool in the debate process to solve the problem of limited and different knowledge backgrounds. Then, we propose an adaptive knowledge selection method to guarantee the accuracy and personalization of knowledge. This method allows agents to choose whether to use external knowledge in each conversation round according to their own needs. 
    Our experimental results on six datasets show that our method achieves state-of-the-art results compared to existing single-agent and multi-agent methods. Further analysis reveals that the introduction of retrieval knowledge can help the agent to break cognitive islands in the debate process and effectively improve the consistency and correctness of the model. Moreover, MADKE using Qwen1.5-72B-Chat surpasses GPT-4 by +1.26\% on average in six datasets, which validates that our method can help open-source LLMs achieve or even surpass the performance of GPT-4.  Our code is available at \url{https://github.com/FutureForMe/MADKE}.
\end{abstract}

%% file: sections/1_introduction.tex
\section{Introduction}   
Multi-expert debates utilize multiple perspectives and critical discussions to facilitate rapid error correction and the integration of diverse knowledge, significantly improving the quality and scientific rigour of decision-making \citep{slonim2021autonomous,ke2024enhancing,chen2023multi}. This promotes interdisciplinary collaboration and the fusion of knowledge across different fields, playing a crucial role in medical consultations, judicial debates, and policy-making~\citep{smit2023we,hua2023war,williams2023epidemic}. Large language models (LLMs) have demonstrated remarkable advantages in various natural language processing (NLP) tasks due to their efficient semantic understanding and language generation capabilities~\citep{openai2023gpt4,touvron2023llama,pu2023summarization}. Employing LLMs as the expert brain in multi-expert debates for dialectical conversations and knowledge integration can simulate real expert dialogues more effectively and ensure extensive knowledge coverage~\citep{xiao2024chainofexperts}. 

\begin{figure}[t]
\centering
\includegraphics[width=\linewidth]{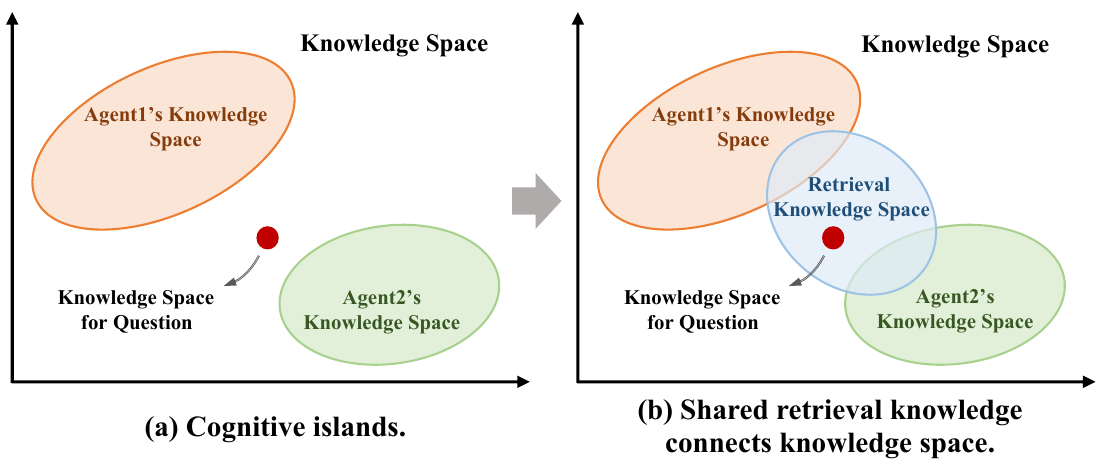}
\vspace{-4mm}
\caption{The knowledge space of agents and required knowledge space for answering questions. (a) Agents have limited and different knowledge spaces, leading to the cognitive islands problem. (b) Constructing a shared knowledge pool by retrieving external knowledge reduces background knowledge differences and connects the knowledge spaces of agents.}
\label{fig_cognitive_island}
\end{figure}

\begin{figure}[t]
\centering
\includegraphics[width=\linewidth]{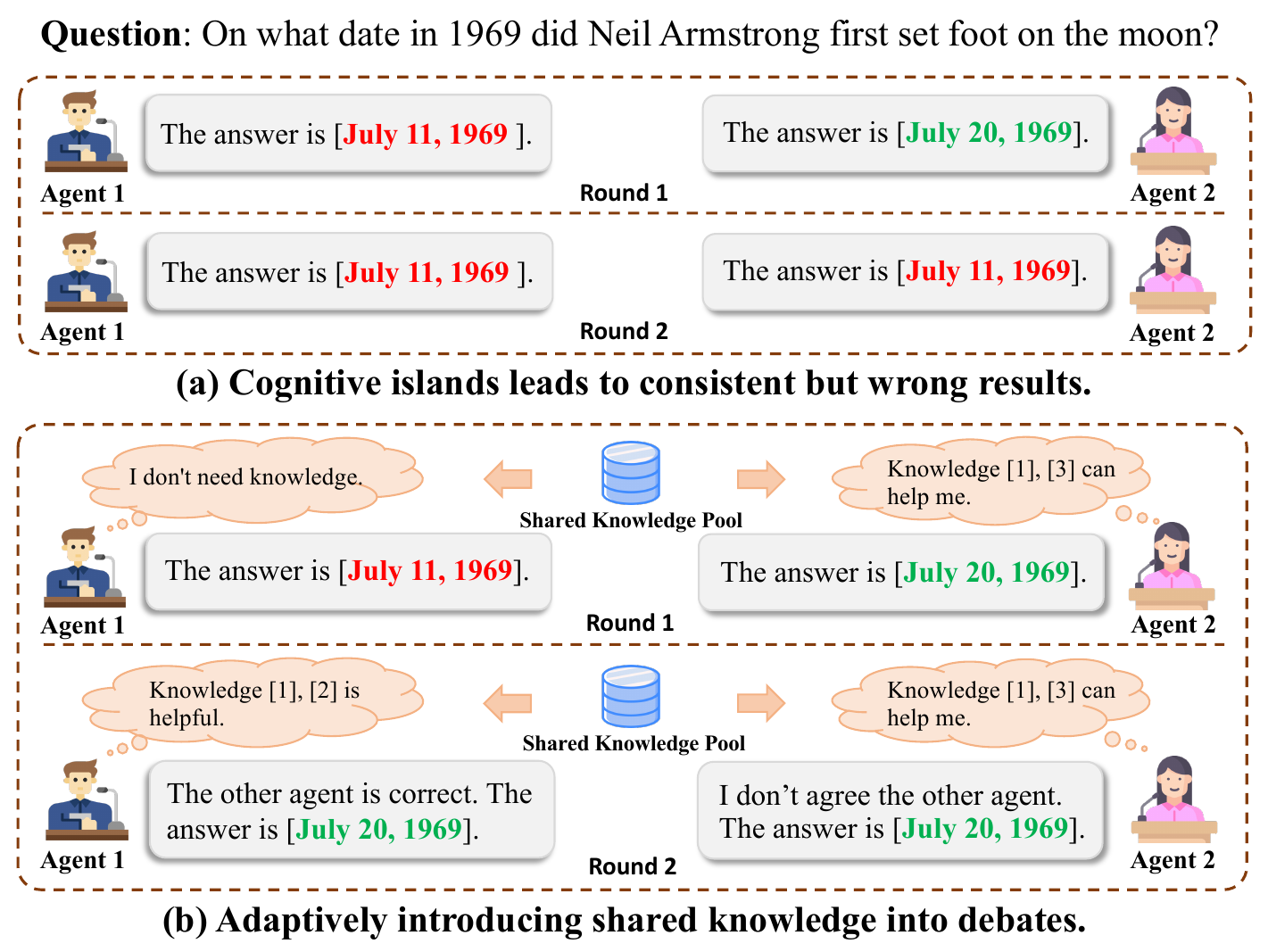}
\vspace{-3mm}
\caption{Comparison on whether to introduce external knowledge. (a) The cognitive islands problem in debates often leads to consistent but wrong outcomes. (b) Adaptively introducing shared knowledge guarantees both accuracy and personalization, alleviating the cognitive islands problem.}
\label{fig_introduction}
\end{figure}

Current research has developed LLM-based multi-agent debate (MAD) systems~\citep{qian2023communicative,wang2024survey}. They utilize LLMs as the decision-making brains of the agents and achieve information integration through collaborative or competitive interactions during discussions, effectively enhancing model consistency~\citep{li2023camel,du2023improving,xiong2023examining,xu2023towards}. 
The success of these works largely relies on the diversity of LLMs, which ensures the provision of varied perspectives and insights, facilitating a comprehensive analysis of problems and avoiding bias, thereby increasing the robustness and reliability of the models~\citep{chen2023chatgpt,keskar2019ctrl}. However, this diversity can also lead to differences in the internal knowledge activated by the models~\citep{xu2024survey}, creating background knowledge disparities among agents and causing cognitive islands issues, as illustrated in Figure \ref{fig_cognitive_island}(a). Cognitive islands refer to the situation where agents, due to limited and differing background knowledge, stubbornly adhere to incorrect viewpoints or easily change correct viewpoints when their knowledge space is insufficient to answer questions, making it difficult to achieve correct and consistent solutions. Specifically, as shown in Figure \ref{fig_introduction}(a), $Agent 1$ initially holds an incorrect viewpoint, while $Agent 2$ is correct; during subsequent debates, $Agent 1$ persists with the incorrect viewpoint, and $Agent 2$ changes its correct viewpoint, resulting in incorrect reasoning outcomes. Additionally, the hallucination~\citep{huang2023survey,yin2023large} of LLMs introduces factual errors in the reasoning process, further exacerbating the impact of cognitive islands. Thus, \textit{breaking the barriers of cognitive islands among multiple agents is crucial for improving the model's cognitive consistency and reasoning performance.}

Retrieval-augmented generation (RAG) provides LLMs with broader knowledge coverage and empirical support by retrieving external knowledge~\citep{izacard2022few,feng2023trends}. Therefore, we propose a novel multi-agent debate with knowledge-enhanced framework (\textbf{MADKE}) to offer agents a shared knowledge platform that reduces background knowledge differences. First, we introduce retrieved external knowledge into the multi-agent debate reasoning to connect the cognitive spaces of different agents and mitigate the cognitive islands problem, as illustrated in Figure \ref{fig_cognitive_island}(b). Further, to avoid misleading decisions caused by noise in retrieved knowledge and the weakening of agent personalization, we propose an adaptive knowledge selection method. This method allows agents to choose whether to use external knowledge based on their needs before expressing their views in each round, thereby ensuring the accuracy and personalization of the knowledge, as shown in Figure \ref{fig_introduction}(b). 
We conducted experiments using different LLMs across six datasets, and the results showed that our model achieved excellent performance, demonstrating strong applicability. Further analysis revealed that introducing external retrieval knowledge effectively mitigates the cognitive islands problem in multi-agent debate systems. Moreover, applying our method to the Qwen1.5-72B-Chat model resulted in an average improvement of +1.26\% over GPT-4 across the six datasets, indicating that our approach can significantly enhance the performance of open-source LLMs to reach the level of more advanced models.

The contributions of this work are as follows:
\begin{itemize}
    \item We proposed a novel multi-agent debate with knowledge-enhanced framework that demonstrated excellent performance across various LLMs, showcasing strong applicability.
    \item We provided a shared knowledge pool for agents in the debate, reducing background knowledge disparities and connecting the knowledge spaces of different agents, effectively alleviating the cognitive islands problem.
    \item We introduced an adaptive knowledge selection method, allowing agents to choose shared knowledge based on their needs before expressing their views, ensuring accuracy and personalization of the knowledge.
    \item Experimental results showed that our method achieved outstanding performance and demonstrated good generalizability. Additionally, this approach helped open-source LLMs reach the performance levels of more advanced models.
\end{itemize}

%% file: sections/2_relatedwork.tex
\section{Related Work}
\subsection{Multi-Agent Collaboration}
The outstanding performance of LLMs~\citep{wei2022emergent} has laid the groundwork for developing autonomous agents~\citep{gravitasauto}. As an essential aspect of autonomous agents, multi-agent collaboration has received widespread attention and is divided into cooperative and adversarial interaction~\citep{xi2023rise}. 
Cooperative interaction requires the construction of agents with different roles, each of which performs its own duties and works together to complete the final task~\citep{li2023camel,zhang2023proagent}. For example, \citet{park2023generative} demonstrated that generative agents could produce believable individual and emergent social behaviours. Further, \citet{qian2023communicative} carefully designed agents with different roles to complete the software development.

Adversarial interactive methods usually build multiple agents and complete reasoning tasks in a debate manner~\citep{fu2023improving,wang2023unleashing,xiong2023examining}. \citet{du2023improving} and \citet{liang2023encouraging} proposed using the form of multi-agent debate to improve the consistency and reasoning capabilities of language models. \citet{chan2023chateval} used the multi-agent debate to evaluate the reply quality of LLMs. 
Benefiting from the diversity generated by language models, existing multi-agent debate methods have achieved notable success. However, this diversity, coupled with the issue of hallucinations, can result in cognitive islands. In contrast to existing approaches, we propose integrating shared knowledge to connect the knowledge spaces between agents, thereby breaking down cognitive islands.

\subsection{Retrieval Augmented Models}

\begin{figure*}[ht]
\centering
\includegraphics[width=\textwidth]{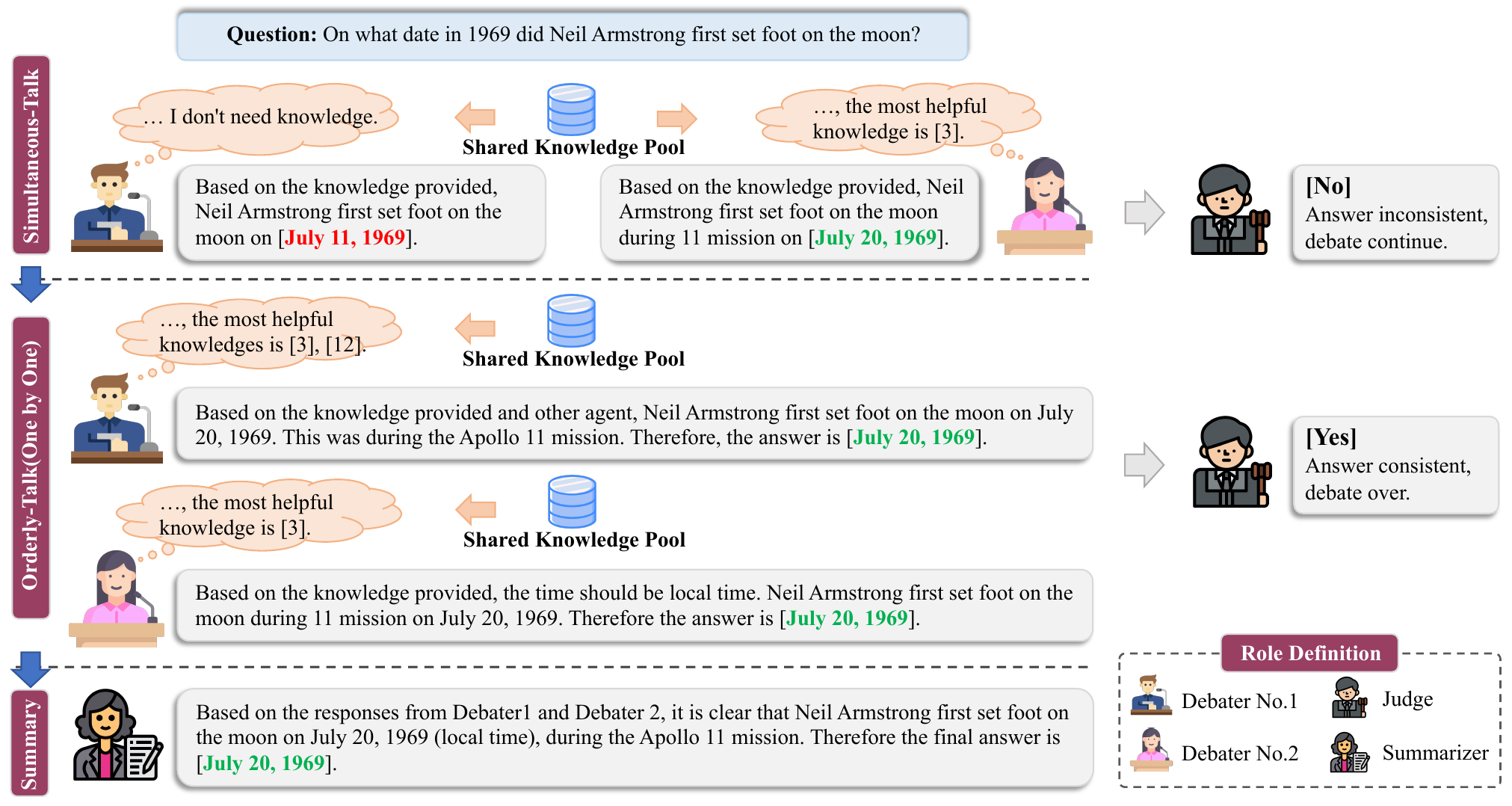}
\vspace{-4mm}
\caption{The framework of our multi-agent debate. Upon receiving the question, debaters initially present their views in the simultaneous-talk stage, followed by orderly individual debates in the next stage. Before making each point, they select relevant knowledge from the shared knowledge pool. A judge then assesses whether to conclude the debate after each round, and a summarizer compiles the final answers post-debate. }
\label{fig_main_framework}
\end{figure*}

Incorporating external knowledge has become a pivotal strategy to fill knowledge gaps in LLMs, enhance their knowledge coverage, and mitigate hallucinations~\citep{shuster2021retrieval,izacard2022few,ZOU2024124306}. Additionally, introducing external knowledge relevant to the query via prompts can improve the model's reasoning capabilities\citep{shi2023replug,ram2023ralm}. \citet{yao2022react} combined reasoning and action in LLMs, guiding them to invoke external retrieval modules via a few-shot approach. To further enhance retrieval capabilities, multiple-round retrieval-enhanced reasoning methods were proposed~\citep{he2022rethinking,shao2023enhancing,jiang2023active,trivedi2022interleaving}. Additionally, some works have utilized search engines for knowledge retrieval, aiding LLMs in complex reasoning tasks~\citep{nakano2021webgpt,liu2023webglm}. 
Compared to existing methods, we fully consider the importance of personalized knowledge in multi-agent debates and propose an adaptive knowledge selection method. This method allows agents to autonomously choose relevant knowledge from a shared knowledge pool, thereby filtering out noisy knowledge and ensuring both accuracy and personalization.

%% file: sections/3_method.tex
\section{Method}

In this work, we propose a multi-agent debate framework based on knowledge enhanced to mitigate the challenges posed by cognitive islands. The procedure is depicted in Algorithm \ref{alg:debate}, with an overview of the methodology illustrated in Figure \ref{fig_main_framework}. We introduce the construction and utilization methods of the shared knowledge pool in \S~\ref{shared_knowledge_pool} and detail the multi-agent debate with knowledge-enhanced framework in \S~\ref{debate_process}.


\subsection{Shared Knowledge Pool} \label{shared_knowledge_pool}
Due to the limitations of the agents' background knowledge and the differences in background knowledge among different agents, agents encounter cognitive islands during debates. Drawing from existing methods of knowledge enhancement, we construct a shared knowledge pool by retrieving knowledge about queries, and allow agents to adaptively select this knowledge to ensure accuracy and personalization of the knowledge.

\subsubsection{Construction of Shared Knowledge Pool}
To ensure the quality and timeliness of retrieved knowledge, we selected Wikipedia and the Google search engine to construct the shared knowledge pool.

\paragraph{\textbf{Wikipedia Retrieval.}} 
To effectively retrieve knowledge from the Wikipedia corpus, we utilize the dense passage retriever (DPR)~\citep{karpukhin2020dense}. DPR employs the twin-tower mechanism for encoding questions and passages, which is both straightforward to implement and yields superior results. In the search results, we retained the top 5 most relevant passages.

\paragraph{\textbf{Google Retrieval.}}
Similar to \citet{luo2023sail}, we input queries into the Google search engine and return up to the top 10 search results. Each result contains fields such as a title, a brief description, page content, and page URL. Considering the length of web page content and potential noise, we add only brief descriptions to the shared knowledge pool to assist agents in selecting the most appropriate knowledge.

Based on the above retrieval method, we can construct a shared knowledge pool for each question $q$, with the specific formula as follows:
\begin{equation}
    SKPool(q) = WR(q) \cup GSE(q)
\end{equation}
where $WR(\cdot)$ represents the Wikipedia retriever and $GSE(\cdot)$ represents the Google search engine retrieval.

\subsubsection{Adaptive Knowledge Selection}

Knowledge retrieved often contains irrelevant noise knowledge, which can negatively impact agents and lead to incorrect reasoning outcomes. Additionally, as agents may produce different viewpoints throughout a debate, supplying identical knowledge for each agent can impede their argumentative thinking. Consequently, we propose an adaptive knowledge selection method, which allows agents to independently identify and select the most beneficial knowledge from the shared knowledge pool before generating their viewpoints. This method filters out noise knowledge and provides personalized knowledge for each agent, ensuring they can engage in effective debates. The specific formula is shown in Eq. (\ref{equ_2}). For further details, see Prompt \ref{prompt_self_selection}.
\begin{align}
    K_i^r = \left\{
    \begin{array}{ll}
        KnoSelect(q, SKPool(q))               & r = 1 \\
        KnoSelect(q, s_i^{(r-1)}, SKPool(q))  & r > 1
    \end{array}
    \right.  \label{equ_2}
\end{align}
where $r$ represents the debate round, $s_i^{(r-1)}$ represents the stance of debater $D_i$ in $r-1$-th round, and $K_i^r$ represents the knowledges chosen by debater $D_i$ before expressing their viewpoint in the $r$-th round.

\input{prompts/adaptive_evidence_selection}

\subsection{Knowledge-Enhanced Multi-Agent Debate}  \label{debate_process}
We propose a knowledge-enhanced multi-agent debate framework with the algorithm outlined in Algorithm \ref{alg:debate}. By introducing shared knowledge, this framework breaks down cognitive islands and connects the knowledge spaces of different agents, effectively improving cognitive consistency and the accuracy of reasoning.

\subsubsection{Role Definition} \label{role_define}
Our debate framework is structured around three key roles: the debater, the judge, and the summarizer. Each role has specific responsibilities, as outlined below. See the Appendix \ref{appendix_role} of role definition prompts. 

\paragraph{\textbf{Debater.}} 
The debater has two primary duties: First, based on the question, the debater must identify and select knowledge from the shared knowledge pool to help them answer it. This process aims to filter out irrelevant or misleading information. Second, the debater needs to generate a new round of responses based on the debate history and selected knowledge.

\paragraph{\textbf{Judge.}}
The responsibilities of the judge are relatively simple yet pivotal. The judge is tasked with determining whether the debaters agree with their responses. If a consensus is reached, the judge is responsible for promptly ending the debate. This action is essential to prevent the debate that extended debates could transform correct responses into incorrect ones.

\paragraph{\textbf{Summarizer.}}
The role of the summarizer is to summarize the answers given by the different debaters to form a final response after the debate. Note that this role demands strict impartiality. We require the summarizer not to have subjective opinions but only to summarize the viewpoints of other debaters.

\input{tables/debate_algorithm}
\subsubsection{Debate Process} 
\paragraph{\textbf{Overall Process.}}
As illustrated in Figure \ref{fig_main_framework}, the debate process is structured into three distinct stages: simultaneous-talk, orderly-talk, and summary. During the simultaneous-talk stage, each debater independently presents their views, uninfluenced by the responses of others. In the orderly-talk stage, debaters, one after another, build upon the views previously expressed. Each debater speaks once per round, and after each round, the judge evaluates whether to conclude the debate. At the end of the debate, the summarizer will summarize the final answer based on the views of all debaters. Notably, before presenting their views, each debater selects relevant knowledge from the shared knowledge pool created by retrieving knowledge from the Wikipedia corpus and Google engine based on the question. 

\paragraph{\textbf{Simultaneous-Talk.}} During the simultaneous-talk, the primary duty of each debater is to independently formulate a response to the question. This response should be as accurate as possible and grounded in the available knowledge of their own selection. The formula is shown in Eq (\ref{equ_3}). The details in Prompt \ref{prompt_first_debate}.   
\begin{equation}
    s_i^r = D_i(q, K_i^r) \label{equ_3}
\end{equation}
where $s_i^r$ represents the stance of debater $D_i$ in the $r$-th round.
\input{prompts/simultaneous_talk}

\paragraph{\textbf{Orderly-Talk.}} In this stage, each debater is expected to consider the responses of the other debaters from the previous round as well as the knowledge of their own choice. This process involves a discussion with the other debaters to get a final answer collaboratively. This stage is critical for integrating diverse viewpoints and knowledge into a cohesive conclusion. The formula is shown in Eq (\ref{equ_4}). The details in Prompt \ref{prompt_second_debate}.
\begin{equation}
    s_i^r = D_i(q, s_1^{(r-1)},...,s_m^{(r-1)}, K_i^r) \label{equ_4}
\end{equation}
where $s_1^{(r-1)},...,s_m^{(r-1)}$ represents the stances of debater $D_i$ and the other debaters in the $(r-1)$-th round.
\input{prompts/orderly_talk}

\paragraph{\textbf{Judge.}} Upon completing each round of discussion, the judge evaluates the debaters' viewpoints to decide whether to conclude the debate. The formula is shown in Eq (\ref{equ_5}). For details, see Prompt \ref{prompt_juedge}.
\begin{equation}
    JudgeResult = J(s_1^r,...,s_m^r) \label{equ_5}
\end{equation}
where $J(\cdot)$ represents the judge agent, $JudgeResult \in \{Yes, No\}$ represents the judgment result. $Yes$ indicates a consensus has been reached, and the debate ends. $No$ indicates no consensus has been reached, and the debate continues.
\input{prompts/judgement}

\paragraph{\textbf{Summary.}} After the end of the debate, the responsibility of the summarizer is to summarize the final answers from all the debaters’ responses. The summarizer must carefully synthesize these varied viewpoints into a coherent and comprehensive conclusion. The formula is shown in Eq (\ref{equ_6}). For details, see Prompt \ref{prompt_summary}.
\begin{equation}
    ReaResult = S(s_1^r,...,s_m^r) \label{equ_6}
\end{equation}
where $S$ represents the summarizer agent, and $ReaResult$ represents the final result of the debate.

\input{prompts/summary}

%% file: prompts/adaptive_evidence_selection.tex
\begin{figure}[h]
\centering
\begin{prompt}[title={Prompt \thetcbcounter: Adaptive Knowledge Selection}, label=prompt_self_selection]
Please select knowledge from the knowledge pool that will help you answer the question. If the knowledge pool does not contain the information needed to answer the question, add [No Found] at the end of your response. If the knowledge pool has knowledge that can help you answer the question, please return up to 3 of the most helpful knowledge. Put the number in square brackets.\\

\{\textcolor{blue}{\textbf{knowledges}}\} \\

\{\textcolor{blue}{\textbf{agent\_history\_answer}}\}    \#Second round and beyond \\

Question: \{\textcolor{blue}{\textbf{question}}\}

Answer: Let's think step by step!
\end{prompt}
\end{figure}

%% file: tables/debate_algorithm.tex
\begin{algorithm}[t]
\small
\begin{algorithmic}[1]
\Require Complex reasoning questions, $q$
\Require Multiple debaters basen on LLM, $\mathcal{D}=\{D_1,...,D_m\}$
\Require Judge based on LLM, $J$
\Require Summarizer based on LLM, $S$
\Require Wikipedia Retriever $WR$
\Require Google Search Engine $GSE$
\State $SKPool(q) = WR(q) \cup GSE(q)$
\For{$r$ in range($MaxDebateRound$)}
    \If{$r$ is 1}
        \For{$D_i$ in $\mathcal{D}$}
            \LState $K_i^r = KnoSelect(q, SKPool(q))$ 
            \LState $s_i^r = D_i(q, K_i^r)$
        \EndFor
    \Else
        \For{$D_i$ in $\mathcal{D}$}
            \LState $K_i^r = KnoSelect(q, s_i^{(r-1)}, SKPool(q))$
            \LState $s_i^r = D_i(q, s_1^{(r-1)}, ..., s_m^{(r-1)}, K_i^r)$
        \EndFor
    \EndIf

    \If{$J(s_1^r, ..., s_m^r)$ is True}
        break
    \EndIf
\EndFor
\LState $ReaResult = S(s_1^r, ..., s_m^r)$
\\ \Return $ReaResult$
\end{algorithmic} 
\caption{Knowledge-Enhanced Multi-Agent Debate}
\label{alg:debate} 
\end{algorithm}

%% file: prompts/simultaneous_talk.tex
\begin{figure}[h]
\centering
\begin{prompt}[title={Prompt \thetcbcounter: Simultaneous-Talk}, label=prompt_first_debate]
Answer the question as accurately as possible based on the information given, and put the answer in the form [answer].
Here is an example:\\
\{\textcolor{blue}{\textbf{example}}\}\\
(END OF EXAMPLE)\\

\{\textcolor{blue}{\textbf{knowledges}}\}\\

Question: \{\textcolor{blue}{\textbf{question}}\} \\
Answer: Let's think step by step!
\end{prompt}
\end{figure}

%% file: prompts/orderly_talk.tex
\begin{figure}[h]
\centering
\begin{prompt}[title={Prompt \thetcbcounter: Orderly-Talk}, label=prompt_second_debate]
There are a few other agents assigned the same task, it's your responsibility to discuss with them and think critically. You can update your answer with other agents' answers or given knowledges as advice, or you can not update your answer. Please put the answer in the form [answer].\\

\{\textcolor{blue}{\textbf{knowledges}}\} \\

\{\textcolor{blue}{\textbf{answer\_from\_other\_agents}}\} \\

\{\textcolor{blue}{\textbf{your\_historical\_answer}}\} \\

Question: \{\textcolor{blue}{\textbf{question}}\} \\
Answer: Let's think step by step! 
\end{prompt}
\end{figure}

%% file: prompts/judgement.tex
\begin{figure}[h]
\centering
\begin{prompt}[title={Prompt \thetcbcounter: Judgement}, label=prompt_juedge]
The answer of the agents are typically denoted with the [answer] format. Your task is to extract each agent's answer and evaluate the consistency of their answers to the question. If all agents have provided correct and consistent answers, respond with [Yes]. If their answers are inconsistent, respond with [No]. Please ensure to encase your response - Yes or No - within square brackets.\\

Question: \{\textcolor{blue}{\textbf{question}}\} \\

Agent Responses: \{\textcolor{blue}{\textbf{all\_answers\_from\_agents}}\} \\

Answer: Let's think step by step!
\end{prompt}
\end{figure}

%% file: prompts/summary.tex
\begin{figure}[h]
\centering
\begin{prompt}[title={Prompt \thetcbcounter: Summary}, label=prompt_summary]

Please summarize the final answer from answer of all agents. Place the final answer of question in the form of [answer].
Here is some examples: \\
\{\textcolor{blue}{\textbf{examples}}\} \\
(END OF EXAMPLE) \\

Question: \{\textcolor{blue}{\textbf{question}}\}  \\

\{\textcolor{blue}{\textbf{all\_answers\_from\_agents}}\}  \\

Answer: Let's think step by step!
\end{prompt}
\end{figure}

%% file: sections/4_experiments.tex

\input{tables/dataset_table}
\section{Experiment}
\subsection{Dataset}

We evaluate the effectiveness of our method on three different types of datasets, including single-hop reasoning: TriviaQA~\citep{joshi2017triviaqa} and NQ~\citep{kwiatkowski2019natural}, multi-hop reasoning: HotpotQA~\citep{yang2018hotpotqa} and 2WikiMQA~\citep{ho2020constructing}, fact verification: FEVER~\citep{thorne2018fever}. Additionally, to further evaluate the model's generalization in vertical medical domains, similar to \citet{xiong2024benchmarking}, we constructed MMLU-Med from six medical subcategories of the MMLU~\citep{hendrycks2021measuring} dataset. Considering time and funding constraints, we randomly sampled some data from each dataset for experiments. The specific dataset statistics are shown in Table \ref{tab_dataset_statis}. Details of the dataset are provided in the Appendix~\ref{appendix_dataset}.

%

\subsection{Baseline}

To compare the performance of our models further, we selected some baselines from single-agent reasoning and multi-agent debate methods and conducted experiments using three backbones: GPT-3.5-Turbo, Qwen1.5-32B-Chat, and Qwen1.5-72B-Chat.
\begin{itemize}
    \item \textbf{Standard Prompting}~\citep{NEURIPS2020_1457c0d6} directly generates the final answer with a one-shot demonstration to standardize the output format.

    \item  \textbf{Chain-of-Thought Prompting}~\citep{kojima2022large} uses the magic prompt "let's think step by step" and one-shot CoT demonstration to generate reasoning steps before the final answer. 

    \item  \textbf{Self Consistency}~\citep{wang2022self} samples a diverse set of reasoning paths and then selects the most consistent answer by marginalizing all possible reasoning paths.

    \item  \textbf{Retrieval-Augmented Generation} uses all knowledge retrieved from the Google search engine and Wikipedia to generate the final answer.


    \item  \textbf{Multi-Agent Debate}~\citep{du2023improving} utilizes multiple agents to propose and debate their individual responses and reasoning processes over multiple rounds to arrive at a final answer.
\end{itemize}

\subsection{Evaluation Metric}
The evaluation metric of existing datasets mainly uses extract match (EM)~\citep{jiang2023active}. We believe that since the answer categories are fixed for discriminative answers, LLM can output the answer in the correct format after given examples in the prompt. Therefore, we use EM to evaluate FEVER and MMLU-Med datasets. For generative answers, EM is more stringent and cannot accurately judge answers better. Therefore, for TriviaQA, NQ, HotpotQA, and 2WikiMQA, in addition to using the F1 metric for evaluation, we also extract answers from the response and use GPT-4~\citep{openai2023gpt4} to evaluate the correctness based on the ground truth answer~\citep{zheng2023judging}, which is called GPT4 Score. 
See the Appendix \ref{appendix_eval} for details.

To further evaluate cognitive islands during the debate process, we used two metrics: the consistent and correct ratio ($Co2$) and the consistency ratio ($Cons$). The specific formulas are shown in Eq. (\ref{equ_7}) - (\ref{equ_8}).
\begin{align}
    &Co2 = \frac{\#\  consistent\ \&\ correct}{ N}  \label{equ_7} \\
    &Cons = \frac{\#\  consistent}{N}  \label{equ_8}
\end{align}
where $N$ represents the total number of data.

\input{tables/main_result}
\subsection{Implementation Details}
We use GPT-3.5-Turbo and GPT-4-Turbo by Azure OpenAI API 2023-07-01-preview\footnote{\url{https://learn.microsoft.com/en-us/azure/ai-services/openai/reference}} to complete our experiments. Additionally, our experiments also used the Qwen1.5-Chat series\footnote{\url{https://huggingface.co/Qwen}} \citep{qwen} of open-source models. In order to ensure the diversity and stability of the debate process, we set the temperature parameter to 0.5. The number of agents in the experiment is 2, and the maximum number of debate rounds is 3. 

Knowledge in the shared knowledge pool is retrieved using the Google search engine and Wikipedia corpus. For the Google search engine, we use the Google SERP provided by Serper\footnote{\url{https://serper.dev/}}. We consider the Wikipedia dump on January 20, 2021, and use the index provided by pyserini\footnote{\url{https://github.com/castorini/pyserini}} for knowledge retrieval. The retrieval numbers of the Google search engine and the Wikipedia corpus are 5 and 5.

%% file: tables/dataset_table.tex
\begin{table*}[]
\centering

\resizebox{\textwidth}{!}{
\begin{tabular}{ccccc}
\toprule
Dataset Name  & Dataset Domain  & Dataset Type & Answer Type  & Samples           \\ \midrule
TriviaQA~\citep{joshi2017triviaqa}          & Open      & Single-Hop Reasoning  & Generative                  &  500 \\
NQ~\citep{kwiatkowski2019natural}           & Open      & Single-Hop Reasoning  & Generative                  &  500 \\
HotpotQA~\citep{yang2018hotpotqa}           & Open      & Multi-Hop Reasoning   & Generative                  &  500  \\
2WikiMQA~\citep{ho2020constructing}         & Open      & Multi-Hop Reasoning   & Generative                  &  500  \\
FEVER~\citep{thorne2018fever}               & Open      & Fact Checking         & Discriminant / 3 categories &  500 \\
MMLU-Med~\citep{hendrycks2021measuring}            & Medicine  & Multi-Choice          & Discriminant / 4 categories &  595 \\
\bottomrule
\end{tabular}
}
\caption{Dataset statistics table.}
\label{tab_dataset_statis}
\end{table*}

%% file: tables/main_result.tex
\begin{table*}[]
\centering

\resizebox{\textwidth}{!}{
\begin{tabular}{ccccccccccc}
\toprule
\multirow{2}{*}{Backbone}  & \multirow{2}{*}{Model} & \multicolumn{2}{c}{TriviaQA} & \multicolumn{2}{c}{NQ}  & \multicolumn{2}{c}{HotpotQA} & \multicolumn{2}{c}{2WikiMQA}  & FEVER  \\ \cmidrule(lr){3-4} \cmidrule(lr){5-6} \cmidrule(lr){7-8} \cmidrule(lr){9-10} \cmidrule(lr){11-11}
                           &                        & F1          & GPT4 Score     & F1     & GPT4 Score     & F1          & GPT4 Score     & F1             & GPT4 Score   & EM            \\ \midrule
\rowcolor[gray]{.92}
GPT-4-Turbo                              & SP   & 0.828          & 0.902          & 0.459          & 0.692          & 0.472          & 0.566          & 0.265          & 0.284          & 0.608          \\ \midrule
\multirow{6}{*}{GPT-3.5-Turbo}          & SP    & \secondc{}{0.761} & 0.778          & 0.390          & 0.532          & 0.331          & 0.384          & 0.190          & 0.210          & 0.548          \\
                                  & CoT        & 0.715          & 0.772          & 0.418          & 0.588          & 0.333          & 0.410          & 0.165          & 0.190          & 0.556          \\
                                  & SC  & 0.756       & 0.818          & 0.431          & 0.622       & 0.342          & 0.408          & 0.175          & 0.206          & 0.560          \\
                                  & RAG                & \firstc{}{0.768} & \secondc{}{0.824} & \secondc{}{0.450} & 0.436    & \firstc{}{0.423} & \secondc{}{0.498} & \secondc{}{0.204} & \secondc{}{0.228} & \firstc{}{0.646} \\
                                  & MAD                & 0.712          & 0.798          & 0.401          & \secondc{}{0.648}          & 0.287          & 0.394          & 0.133          & 0.186          & 0.548          \\
                                  & MADKE(Our)         & 0.728          & \firstc{}{0.830}  & \firstc{}{0.453} & \firstc{}{0.736} & \secondc{}{0.405}  & \firstc{}{0.542} & \firstc{}{0.205} & \firstc{}{0.274} & \secondc{}{0.612}          \\ \midrule
\multirow{6}{*}{Qwen1.5-32B-Chat} & SP    & 0.581          & 0.632          & 0.295          & 0.462          & 0.311          & 0.384          & 0.207          & 0.22           & 0.536          \\
                                  & CoT         & 0.587          & 0.644          & 0.276          & 0.452          & 0.315          & 0.382          & 0.220           & 0.228          & 0.546          \\
                                  & SC   & 0.583          & 0.628          & 0.276          & 0.442          & 0.312          & 0.386          & 0.203          & 0.222          & 0.540          \\
                                  & RAG                & 0.446          & 0.554          & 0.269          & 0.436          & 0.309          & 0.420          & 0.183          & 0.248          & \firstc{}{0.664} \\
                                  & MAD                & \secondc{}{0.630}          & \secondc{}{0.704}          & \secondc{}{0.321}          & \secondc{}{0.566}          & \secondc{}{0.363}          & \secondc{}{0.482}          & \secondc{}{0.256}          & \secondc{}{0.286}          & 0.520           \\
                                  & MADKE(Our)     & \firstc{}{0.763} & \firstc{}{0.838} & \firstc{}{0.475} & \firstc{}{0.724} & \firstc{}{0.450}  & \firstc{}{0.570}  & \firstc{}{0.258} & \firstc{}{0.314} & \secondc{}{0.646}          \\ \midrule
\multirow{6}{*}{Qwen1.5-72B-Chat}    & SP        & 0.665          & 0.706          & 0.378          & 0.550           & 0.332          & 0.428          & 0.226          & 0.26           & 0.596          \\
                                  & CoT             & \secondc{}0.715          & 0.776          & \secondc{}{0.406}          & 0.648          & 0.431          & 0.514          & \secondc{}{0.275}          & 0.324          & 0.625          \\
                                  & SC       & 0.659          & 0.700            & 0.369          & 0.548          & 0.336          & 0.430           & 0.207          & 0.242          & 0.596          \\
                                  & RAG          & 0.662          & 0.762          & 0.333          & \secondc{}0.710           & \secondc{}0.438          & 0.538          & 0.217          & 0.304          & \secondc{}{0.668}          \\
                                  & MAD             & 0.703          & \secondc{}0.796          & 0.376          & 0.660           & 0.397          & \secondc{}0.548          & 0.274          & \secondc{}0.344          & 0.516          \\
                                  & MADKE(Our)     & \firstc{}0.787 & \firstc{}0.874 & \firstc{}0.479 & \firstc{}0.748 & \firstc{}0.495 & \firstc{}0.622 & \firstc{}0.284 & \firstc{}0.350  & \firstc{}0.686 \\
\bottomrule
\end{tabular}
}
\caption{Comparative experimental results on all open-domain datasets. Our method surpasses GPT4 on specific datasets. Excluding GPT4's results, for each column of different backbone, the highest, the second figures are highlighted by \setlength{\fboxsep}{0pt}\colorbox[RGB]{ 171, 235, 198 }{green} and \setlength{\fboxsep}{0pt}\colorbox[RGB]{ 253, 235, 208}{orange} backgrounds. All the numbers are presented in \%, and the total score is 100\%.} 
\label{tab_main_result}
\end{table*}

%% file: sections/5_results.tex
\section{Experiment Results}

\subsection{Main Results}
The main results of our method and baseline methods are reported in Table~\ref{tab_main_result}. Our model achieved competitive experimental results on five open-domain datasets, proving the effectiveness of integrating external prior knowledge in enhancing complex reasoning capabilities. Besides, based on the results, we have the following key findings:

\input{tables/cognitive_conflicts}

\textbf{(i) Our model achieved superior performance across all backbone settings compared to single-agent and multi-agent baseline models.} Our MADKE framework, built on Qwen1.5-72B-Chat, shows an average increase of +4.6\% in all metrics compared to the state-of-the-art models across five open-domain datasets. Specifically, on the TriviaQA and HotpotQA datasets, the GPT-4 Score increased by +7.8\% and +7.4\%, respectively. This demonstrates that our multi-agent collaborative reasoning with the introduction of retrieved knowledge has strong adaptability and effectiveness. By incorporating high-quality knowledge and multi-perspective reasoning analysis during the debate process, our approach can more accurately address complex problems, significantly enhancing the model's performance. Additionally, among all the backbones, Qwen1.5-72B-Chat performed the best, indicating that open-source models are gradually catching up with closed-source models, progressively revealing their strong potential and competitiveness.

\textbf{(ii) Introducing shared knowledge can effectively alleviate cognitive islands in debates.} We used the Co2 and the Cons to measure the mitigation of cognitive islands during debates, and the experimental results are shown in Table \ref{tab:cognitive_conflicts}. The experimental results show a slight decrease in the Cons metric after introducing shared knowledge. For example, the MADKE framework based on Qwen-1.5-Chat showed an average reduction of -1.6\% across five datasets compared to the version without knowledge. Introducing personalized knowledge in the debate process enriches and broadens the discussions, which can slightly affect consistency. However, the Co2 metric increased by an average of +7.8\% compared to the version without knowledge. The substantial improvement in the Co2 metric far outweighs the minor impact on the Cons metric. This further demonstrates that providing high-quality and personalized knowledge to debaters helps them better understand and interpret opposing viewpoints, effectively mitigating cognitive islands and enhancing the overall quality of the debate.

\textbf{(iii) Our method has outperformed GPT-4 on specific datasets.} As shown in Table~\ref{tab_main_result}, our approach based on the Qwen1.5-72B-Chat model outperformed GPT-4 in the GPT-4 Score metric by +5.6\%, +5.6\%, +6.6\%, and +7.8\% on the NQ, HotpotQA, 2WikiMQA, and FEVER datasets, respectively. This demonstrates that organizing multiple agents in a debate format for collaborative reasoning, combined with the introduction of accurate and personalized prior knowledge during the debate, can bridge the gap between smaller parameter models and GPT-4, and even surpass GPT-4's performance. This not only validates the effectiveness of the multi-agent debate mechanism but also provides new insights for the future application of smaller parameter models across a broader range of fields.

\subsection{Ablation Study}
We considered the effect of different retrieval sources and the adaptive knowledge selection module on model performance and conducted an ablation study on all open-domain datasets to verify their effectiveness.


\subsubsection{Effect of Retrieve Source}


\begin{figure*}[h]
\centering
\includegraphics[width=\textwidth]{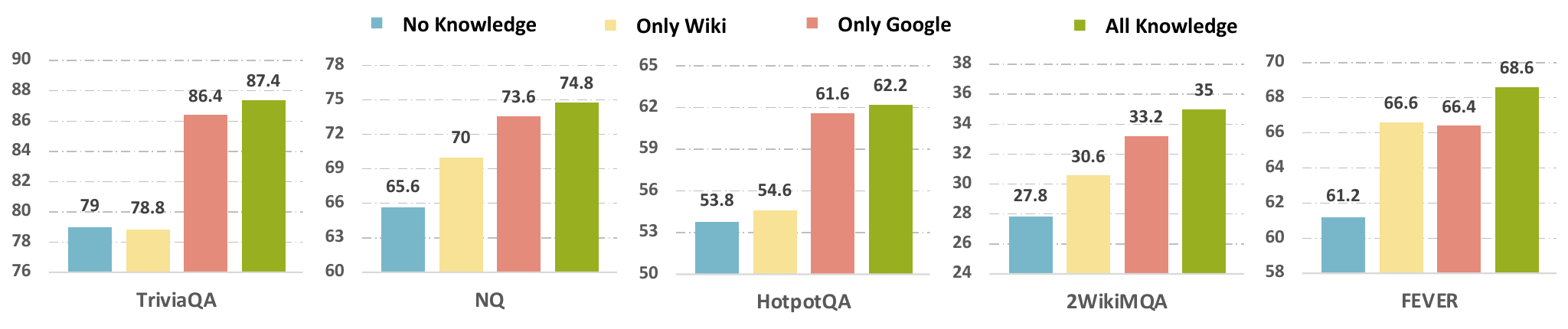}
\vspace{-5mm}
\caption{The performance (EM for FEVER, GPT4 Score for others) of the MADKE model with the Qwen1.5-72B-Chat backbone under different retrieval settings. No Knowledge, Only Wiki, Only Google, and All Knowledge, respectively, indicate not using retrieval knowledge, using knowledge from the Wikipedia knowledge base, using knowledge from the Google search engine, and using knowledge from both sources.}
\label{fig_retrieval_ablation}
\end{figure*}

\begin{figure*}[h]
\centering
\includegraphics[width=0.9\textwidth]{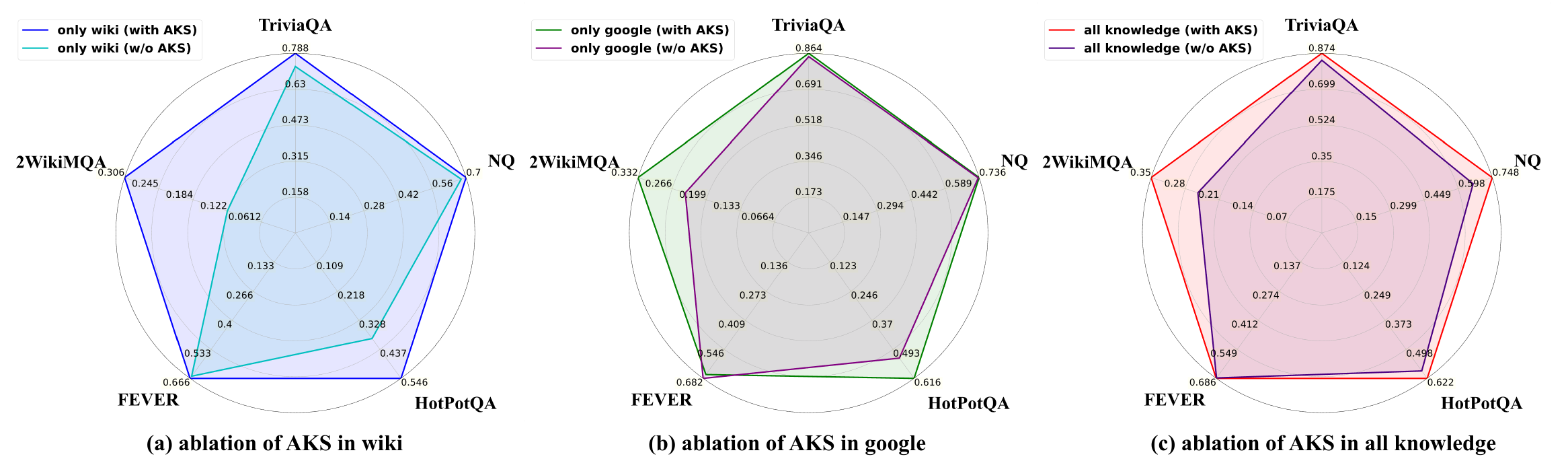}
\caption{The ablation experiment results (EM for FEVER, GPT4 Score for others) of the Adaptive Knowledge Selection module based on Qwen1.5-72B-Chat model under different retrieval knowledge settings. with AKS and w/o AKS indicate using and not using the Adaptive Knowledge Selection module, respectively.}
\label{fig_self_selection}
\end{figure*}

We conducted ablation studies under four settings: without using retrieval knowledge, using only Wikipedia retrieval knowledge, using only Google search engine retrieval knowledge, and using a combination of both. The experimental results are shown in Figure \ref{fig_retrieval_ablation}.

Experimental results indicate that introducing Wikipedia knowledge, Google search knowledge, and both sources of knowledge led to an average improvement of +2.7\%, +6.8\%, and +8.1\% across five datasets compared to not introducing any knowledge. Furthermore, the benefits of using Google search engine retrieval are more pronounced than using the Wikipedia knowledge base. This is because the Google search engine inherently has sorting and filtering functions, resulting in higher relevance and accuracy of the retrieval results. In contrast, the Wikipedia knowledge base stores Wikipedia page texts by cutting them into segments of a certain length, so the retrieval results often contain knowledge related to the question but not directly answering it. This leads to lower benefits from Wikipedia retrieval knowledge, although it is still overall better than not using any retrieval knowledge.

Additionally, the experimental results show that combining retrieval knowledge from both the Google search engine and Wikipedia yields the best results. This is because the combined knowledge contains more comprehensive knowledge, allowing the model to better select useful knowledge to assist in reasoning. In summary, the introduction and combination of different retrieval knowledge significantly enhance the model's reasoning performance.


\begin{figure*}[h]
\centering
\includegraphics[width=\textwidth]{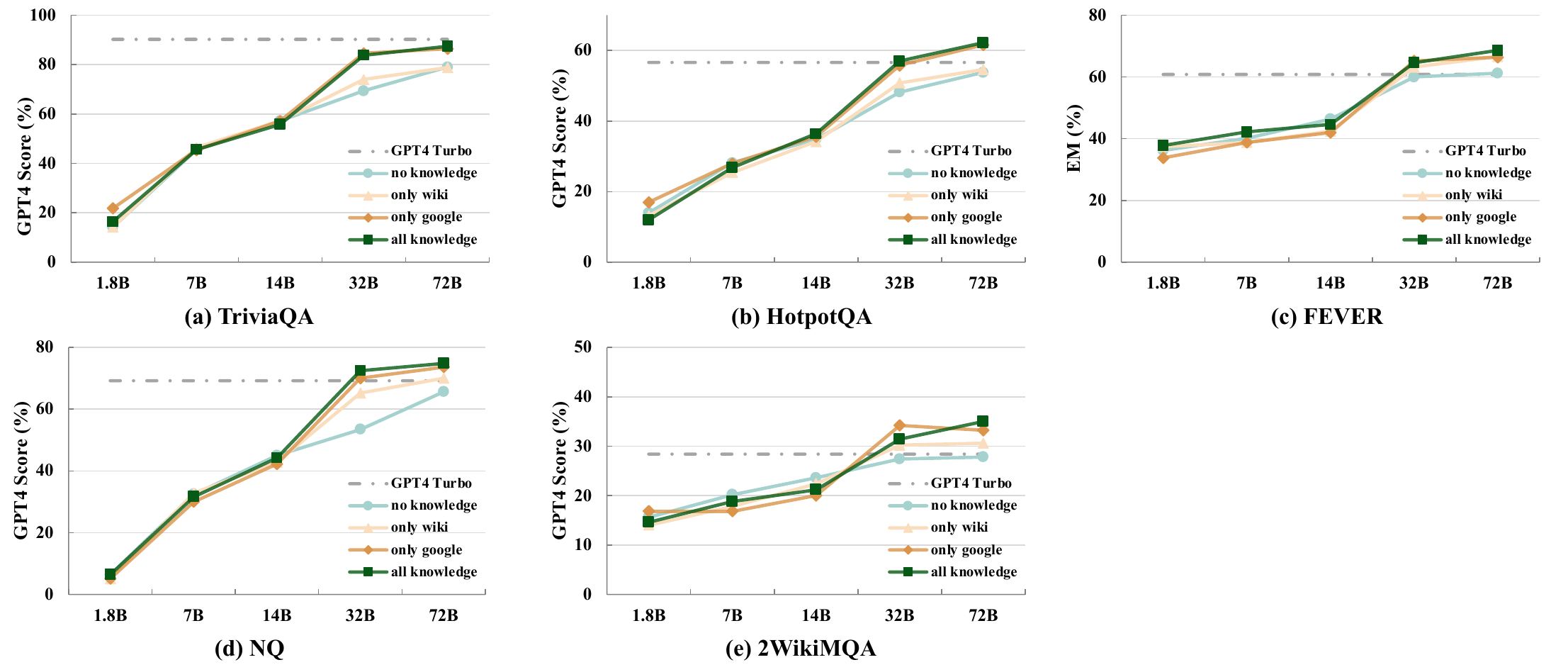}
\vspace{-4mm}
\caption{The performance of Qwen1.5-Chat series models with different parameter scales (1.8B, 7B, 14B, 32B and 72B) on all open-domain datasets using different retrieval knowledge. }
\label{fig_scale_law}
\end{figure*}

\subsubsection{Importance of Adaptive Knowledge Selection}

To further investigate the importance of the adaptive knowledge selection module, we conducted an ablation study based on Qwen1.5-72B-Chat across five open-domain datasets using three retrieval settings. The experimental results are shown in Figure \ref{fig_self_selection}.

The experimental results indicate that under all three retrieval settings, the adaptive knowledge selection module plays a crucial role in improving model performance. This module filters out noise knowledge and provides personalized knowledge to the debating agents before they express their viewpoints, effectively enhancing the consistency and accuracy of the debate and, to some extent, mitigating cognitive islands. Specifically, for multi-hop datasets like 2WikiMQA and HotpotQA, where retrieval results rarely contain knowledge that directly supports answering the questions and often includes much noise, the adaptive knowledge selection module has a more substantial impact, leading to significant performance differences before and after ablation. Additionally, the results also demonstrate that this module has a positive effect on other datasets, further proving its importance in enhancing the overall performance of the model.


\subsection{Effect of Model Parameter Scale}

To evaluate the impact of model parameters, we conducted experiments using five different parameter scales of the Qwen1.5-Chat model on five open-domain datasets. The experimental results are shown in Figure \ref{fig_scale_law}.

From the experimental results, it can be observed that the model performance improves as the model parameter scale increases, validating the positive correlation between parameter scale and performance and further supporting the conclusions of the scaling law~\citep{kaplan2020scaling}. By comparing our model with GPT-4's experimental results, we found that by adaptively introducing prior knowledge in the debate, our method can achieve or even surpass GPT-4's performance when the parameter scale reaches 32B.

Moreover, when the model parameters are 1.8B, 7B, and 14B, prior knowledge does not significantly enhance model performance, indicating that prior knowledge does not have a notable impact on debates in smaller models. However, as the model parameters increase, the performance improvement brought by introducing prior knowledge becomes more apparent. This suggests that larger models have a clearer self-awareness, and when the model parameters exceed 32B, agents can accurately filter out noise data from the retrieved knowledge and select personalized knowledge that is beneficial to them.
By increasing the number of parameters in the debate agents and introducing prior knowledge, the model's self-awareness can be further enhanced, effectively mitigating the issue of cognitive islands and accelerating the model's application.

\subsection{Parameter Sensitivity Study}
To explore the impact of the number of agents and the maximum debate rounds on model performance, we conducted a parameter sensitivity study based on GPT-3.5 Turbo using the TriviaQA and HotpotQA datasets. When investigating the number of agents, we set the maximum debate rounds to 3; when examining the maximum debate rounds, we set the number of agents to 2.

\begin{figure}[h]
\centering
\includegraphics[width=\linewidth]{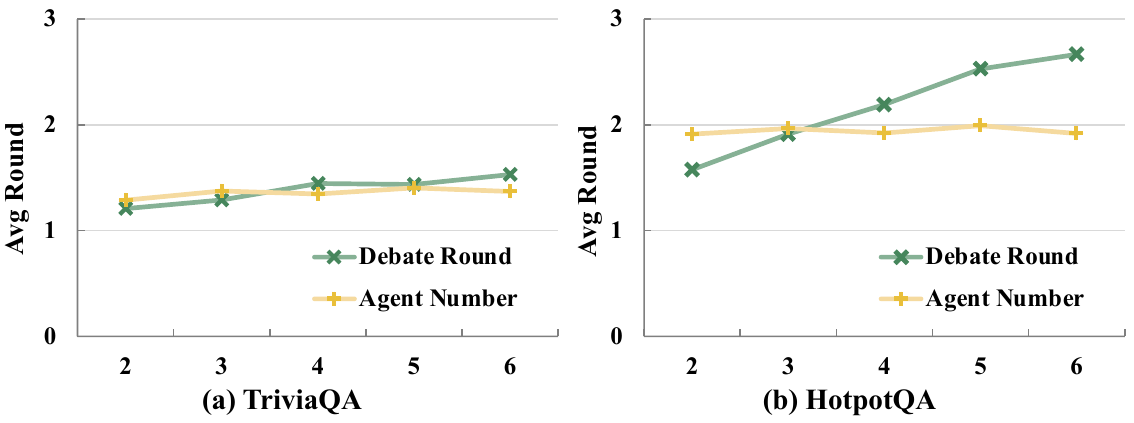}
\vspace{-4mm}
\caption{Effect of the agent number and maximum debate rounds on the average debate round.}
\label{fig_avg_round}
\end{figure}

\begin{figure}[h]
\centering
\includegraphics[width=\linewidth]{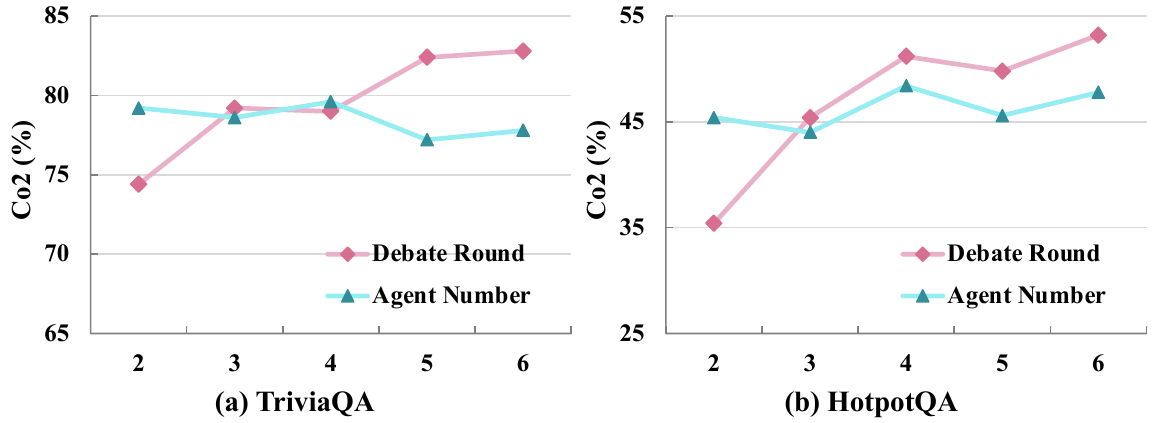}
\vspace{-4mm}
\caption{The performance of different agent number and maximum debate rounds on the Co2 metric.}
\label{fig_co2}
\end{figure}

\begin{figure}[h]
\centering
\includegraphics[width=\linewidth]{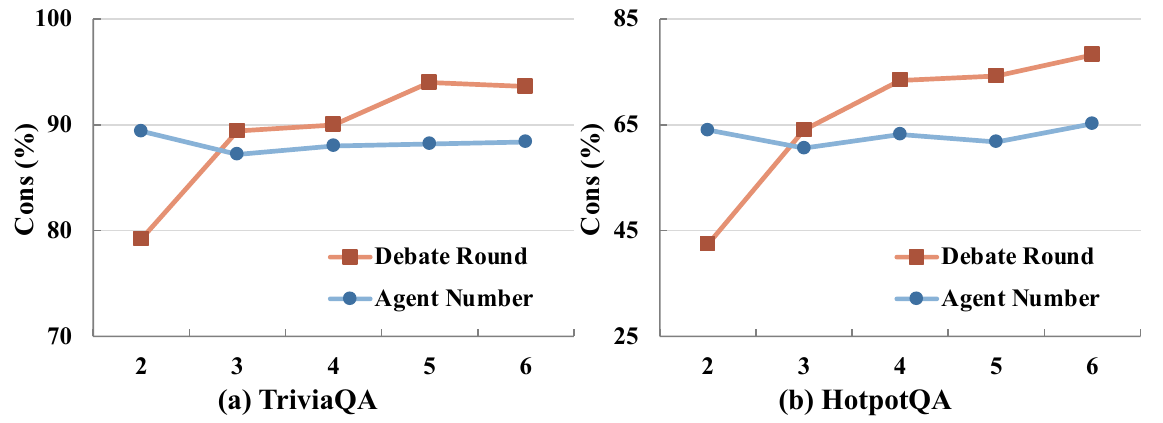}
\vspace{-4mm}
\caption{The performance of different agent number and maximum debate rounds on the Cons metric.}
\label{fig_cons}
\end{figure}



We first studied the impact of the number of agents and the maximum debate rounds on the average debate rounds, with the experimental results shown in Figure \ref{fig_avg_round}. The results indicate that as the maximum debate rounds increase, the average debate rounds also increase. The increase is more significant on multi-hop datasets compared to single-hop datasets. This is because multi-hop reasoning datasets are more challenging, and with more debate rounds, agents continue to debate unresolved issues, increasing in the average debate rounds. 
Subsequently, we explored the impact of the number of agents and the maximum debate rounds on the Co2 and Cons metrics, with the experimental results shown in Figures \ref{fig_co2} and \ref{fig_cons}. The results demonstrate that as the debate rounds increase, both the Cons and Co2 of the debates improve. This suggests that for more complex problems, increasing the number of debate rounds enables the model to reach more accurate and consistent conclusions, thereby alleviating cognitive islands during the debates.

From the experimental results, it can be seen that increasing the number of agents did not significantly help with the average debate rounds, Co2, and Cons metrics. Analyzing the experimental results reveals that as the number of agents increases, debaters in the sequential debate process tend to agree with the more frequently presented answers, reducing the independence and diversity of the debate process. This lack of diversity prevents a significant improvement in the model's reasoning performance. The experimental analysis indicates that our model framework is more suitable for real-world two-sided debate scenarios. In this setting, agents can focus more on high-quality knowledge exchange and effective argumentation, thereby better-enhancing reasoning performance and the reliability of the results. By reducing the number of debaters, the model can more efficiently process information, avoiding redundancy and noise and improving the consistency and correctness of the debate outcomes. Therefore, a two-sided debate setting can better leverage the model's potential, leading to excellent performance in practical applications.

\subsection{Generalizability in Vertical Medical Domain}
\input{tables/med_result_table}
To further evaluate the model's performance on vertical domain datasets, we extracted a subset of medical-related data from the MMLU dataset, named MMLU-Med, to assess our method. Table \ref{tab_med_result} shows the experimental results of GPT-3.5 Turbo and Qwen1.5-72B-Chat. Note that to obtain more relevant medical knowledge, we replaced the Google search engine with a medical textbooks~\citep{jin2021disease} for retrieval.

 We found that our model achieved the best results compared to the baseline model, indicating that our model has strong generalization capabilities in vertical domains and can be quickly transferred to other vertical fields in the future. However, our method still shows a significant performance gap compared to GPT-4. This is primarily because retrieving knowledge in vertical domains is more challenging than in open domains, resulting in lower relevance and higher noise in the retrieved knowledge. The performance drop in single-agent models with knowledge integration (RAG) compared to models without knowledge integration (SP) further supports this point. Therefore, although our model has shown improvements over the baseline, the extent of improvement is not significant. In future research, we will focus on improving the accuracy of vertical domain knowledge retrieval to advance the application of our model in vertical fields.

\subsection{Token Consumption of Different Models}

\begin{figure}[h]
\centering
\includegraphics[width=0.9\linewidth]{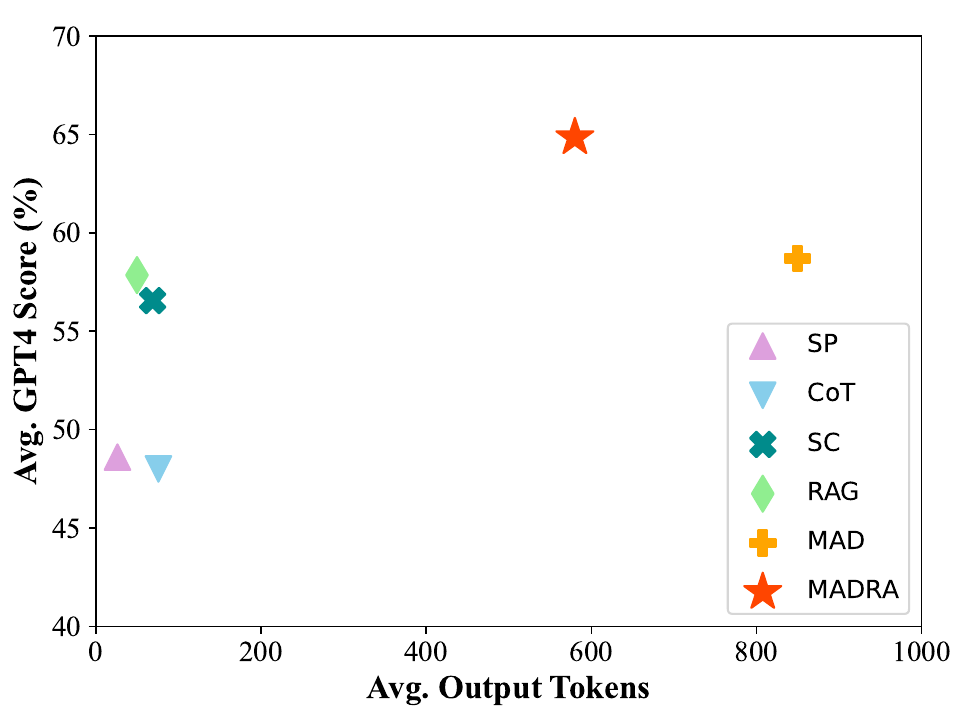}
\caption{Output token consumption and performance (GPT4 Score). Results are averaged over five open-domain datasets.}
\label{fig_token_consumption}
\end{figure}

In the computational cost of large models, the cost of output tokens is greater than that of input tokens. Therefore, we analyzed the output token costs of our model and the baseline methods based on the Qwen1.5-72B-Chat backbone, as shown in Figure \ref{fig_token_consumption}.

From the experimental results, our model incurs higher computational costs than single-agent models, but it also achieves significant performance improvements, which justify these costs. Additionally, we observed that our model reduced computational overhead by -27.5\% compared to the multi-agent debate method (MAD)~\citep{du2023improving}, while improving performance by +6.2\%. This improvement is due to introducing a judge agent, which assesses whether the debaters have reached a consensus and terminates the debate early if they have, thereby significantly reducing the cost of output tokens. Unlike the MAD method, which ends the debate based on a maximum debate round limit, our approach with the judge agent leads to more efficient debates. By incorporating external prior knowledge, we further enhance the consistency and correctness of the debate outcomes, achieving better performance.

\input{tables/case_study_table}
\subsection{Case Study}
To analyze the performance of agents during the debate process, we randomly selected a case from the HotpotQA dataset for analysis, as shown in Table \ref{table_debate_case}. In the first round of viewpoint expression, Debater 1, due to unclear self-awareness, did not use external knowledge and produced an incorrect reasoning answer. In the next round of debate, Debater 1 selected knowledge related to their own viewpoint (Screaming Trees) based on their previous stance (Knowledge [8]), as well as knowledge related to the correct answer (Knowledge [0]). By comparing this retrieved knowledge and referencing the viewpoints of other debaters, Debater 1 corrected their erroneous stance and ultimately reached a consistent and correct answer. This case demonstrates the advantages of introducing prior knowledge, which can alleviate cognitive islands during debates, enhance the overall performance of the model, and help the model handle more complex reasoning scenarios.

%% file: tables/cognitive_conflicts.tex

\begin{table*}[]
\centering

\resizebox{\textwidth}{!}{
\begin{tabular}{cccccccccccc}
\toprule
\multirow{2}{*}{Backbone}         & \multirow{2}{*}{Model} & \multicolumn{2}{c}{TriviaQA}                & \multicolumn{2}{c}{NQ}                      & \multicolumn{2}{c}{HotpotQA}                & \multicolumn{2}{c}{2WikiMQA}                & \multicolumn{2}{c}{FEVER}                   \\ \cmidrule(lr){3-4} \cmidrule(lr){5-6} \cmidrule(lr){7-8} \cmidrule(lr){9-10} \cmidrule(lr){11-12}
                                  &                        & Co2      & Cons     & Co2     & Cons    & Co2     & Cons     & Co2     & Cons    & Co2      & Cons       \\  \midrule
\multirow{4}{*}{GPT-3.5-Turbo}     & w/o knowledge           & \firstc{}0.800    & \firstc{}0.922    & 0.592   & \secondc{}0.818   & 0.432   & \firstc{}0.648    & 0.238   & \firstc{}0.558   & \secondc{}0.312    & \secondc{}0.442      \\
                                  & w wiki knowledge        & 0.748    & 0.842    & \secondc{}0.614	 & 0.754   & 0.432   & 0.586 	& \secondc{}0.248   & 0.492 	& 0.304    & 0.396       \\ 
                                  & w google knowledge      & 0.784 	& 0.852    & 0.58	 & 0.738   & \firstc{}0.470 	 & 0.602 	& \firstc{}0.250   & 0.486 	& 0.294    & 0.394       \\ 
                                  & w all knowledge         & \secondc{}0.792    & \secondc{}0.894    & \firstc{}0.624   & \firstc{}0.840   & \secondc{}0.454   & \secondc{}0.640    & 0.210   & \secondc{}0.498   & \firstc{}0.612    & \firstc{}0.794      \\ \midrule
\multirow{4}{*}{Qwen1.5-32B-Chat} & w/o knowledge           & 0.686    & 0.958    & 0.522   & 0.944   & 0.474   & 0.930    & 0.256   & 0.832   & 0.600    & \secondc{}0.982      \\
                                  & w wiki knowledge        & 0.728 	& 0.954    & 0.642 	 & 0.958   & 0.504 	 & 0.934 	& 0.268   & 0.836 	& 0.622    & \firstc{}0.984       \\ 
                                  & w google knowledge      & \firstc{}0.838 	& \firstc{}0.984    & \secondc{}0.696 	 & \secondc{}0.972   & \secondc{}0.546 	 & \secondc{}0.942 	& \firstc{}0.322   & \secondc{}0.840 	& \firstc{}0.640    & 0.980       \\ 
                                  & w all knowledge         & \secondc{}0.830    & \secondc{}0.966    & \firstc{}0.718   & \firstc{}0.978   & \firstc{}0.564   & \firstc{}0.948    & \secondc{}0.296   & \firstc{}0.868   & \secondc{}0.638    & \firstc{}0.984      \\\midrule
\multirow{4}{*}{Qwen1.5-72B-Chat} & w/o knowledge           & 0.788    & \firstc{}0.996    & 0.652   & \firstc{}0.990   & 0.536   & \firstc{}0.978    & 0.272   & \firstc{}0.966   & 0.612    & \firstc{}0.966      \\
                                  & w wiki knowledge                & 0.784 	& 0.986    & 0.698 	 & 0.988   & 0.536 	 & \secondc{}0.970 	& 0.288   & \secondc{}0.946 	& \secondc{}0.638    & \secondc{}0.960        \\
                                  & w google knowledge              & \secondc{}0.858 	& 0.990    & \secondc{}0.734 	 & \secondc{}0.988   & \secondc{}0.612 	 & 0.966 	& \secondc{}0.308   & 0.944 	& 0.622    & 0.944        \\
                                  & w all knowledge                 & \firstc{}0.872    & \secondc{}0.986    & \firstc{}0.738   & 0.984   & \firstc{}0.612   & 0.958    & \firstc{}0.342   & 0.932   & \firstc{}0.686    & 0.954       \\
\bottomrule
\end{tabular}
}
\caption{The experimental results of our MADRA model under different retrieval knowledge for the Co2 and Cons metrics. w/o and w indicate without and with knowledge. For each column of different backbone, the highest, the second figures are highlighted by \setlength{\fboxsep}{0pt}\colorbox[RGB]{ 171, 235, 198 }{green} and \setlength{\fboxsep}{0pt}\colorbox[RGB]{ 253, 235, 208}{orange} backgrounds. All the numbers are presented in \% and the full score is 100\%.}
\label{tab:cognitive_conflicts}
\end{table*}

%% file: tables/med_result_table.tex
\begin{table}[h]
\centering

\label{tab_med_result}
\resizebox{0.8\linewidth}{!}{
\begin{tabular}{ccc}
\toprule
Backbone                      & Model                  &  EM        \\  \midrule
\rowcolor[gray]{.92}GPT-4-Turbo                    & SP                     & 0.861                \\ \midrule
\multirow{6}{*}{GPT-3.5-Turbo} & SP                     & 0.655                \\
                              & CoT                    & \secondc{}0.681                \\
                              & SC                     & 0.655               \\
                              & RAG                    & 0.545                \\
                              & MAD                    & 0.622               \\
                              & MADRA(Our)             & \firstc{}0.697                \\ \midrule
\multirow{6}{*}{Qwen1.5-72B-Chat}  & SP                & 0.659                   \\
                              & CoT                    & 0.640                  \\
                              & SC                     & 0.671                      \\
                              & RAG                    & 0.524                      \\
                              & MAD                    & \secondc{}0.671                      \\
                              & MADRA(Our)             & \firstc{}0.709                    \\ \bottomrule
\end{tabular}
}
\caption{The experimental results on the MMLU-Med dataset in the medical domain. Excluding GPT4's results, for different backbone, the highest, the second figures are highlighted by \setlength{\fboxsep}{0pt}\colorbox[RGB]{ 171, 235, 198 }{green} and \setlength{\fboxsep}{0pt}\colorbox[RGB]{ 253, 235, 208}{orange} backgrounds.} 
\end{table}

%% file: tables/case_study_table.tex
\begin{table*}[ht]
\small
\centering

\begin{tabular}{p{8cm}p{8cm}}
\toprule
\multicolumn{2}{l}{\textbf{Question}: Which band, Letters to Cleo or Screaming Trees, had more members? }  \\
\multicolumn{2}{l}{\textbf{Answer}: Letters to Cleo}  \\
\midrule
\rowcolor[gray]{.92}
\begin{minipage}[b]{0.08\columnwidth}
    \centering
    \raisebox{-.3\height}{\includegraphics[width=\linewidth]{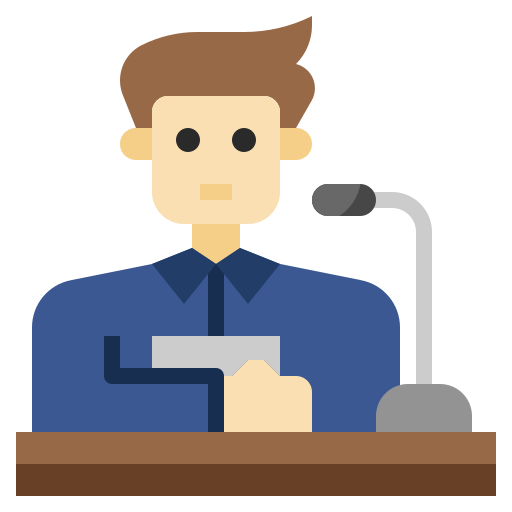}}
\end{minipage}
... On the other hand, the knowledge pool does not provide specific information about the number of members in Screaming Trees. Therefore, we cannot definitively determine if Letters to Cleo or Screaming Trees had more members. [No Found]  & 
\begin{minipage}[b]{0.08\columnwidth}
    \centering
    \raisebox{-.3\height}{\includegraphics[width=\linewidth]{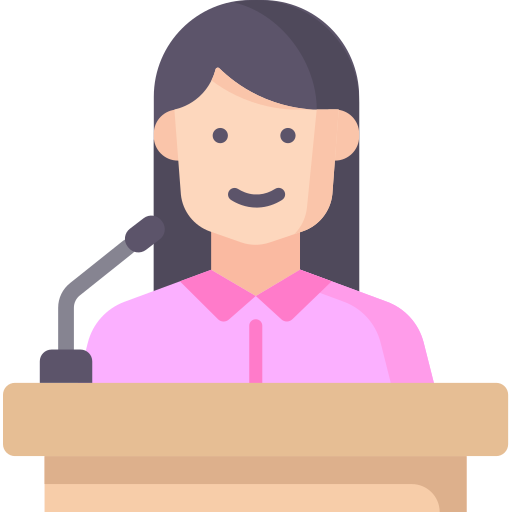}}
\end{minipage}
... [1] "Letters to Cleo" - The knowledge provides information about the band's members, including Kay Hanley, Greg McKenna, Michael Eisenstein, ...[3] "Screaming Trees" - The knowledge states that Screaming Trees had four members: Mark Lanegan, Gary Lee Conner, ... \\

\begin{minipage}[b]{0.08\columnwidth}
    \centering
    \raisebox{-.3\height}{\includegraphics[width=\linewidth]{figures/imgs/debater1.png}}
\end{minipage}
 First, let's compare the number of members in each band. Letters to Cleo is a band that typically consists of four members, while Screaming Trees is a band that typically consists of five members. Based on this information, Screaming Trees had more members than Letters to Cleo. Therefore, the answer is \textcolor{red}{\textbf{[Screaming Trees]}}.  & 
\begin{minipage}[b]{0.08\columnwidth}
    \centering
    \raisebox{-.3\height}{\includegraphics[width=\linewidth]{figures/imgs/debater2.png}}
\end{minipage}
     First, let's gather the information about the bands. Letters to Cleo is an alternative rock band from Boston, Massachusetts. The band's members are Kay Hanley, ...  Based on the information provided, Letters to Cleo had more members compared to Screaming Trees. Therefore, the answer is \textcolor{MyGreen}{\textbf{[Letters to Cleo]}}. \\

\multicolumn{2}{p{16cm}}{\begin{minipage}[b]{0.08\columnwidth}
    \centering
    \raisebox{-.3\height}{\includegraphics[width=\linewidth]{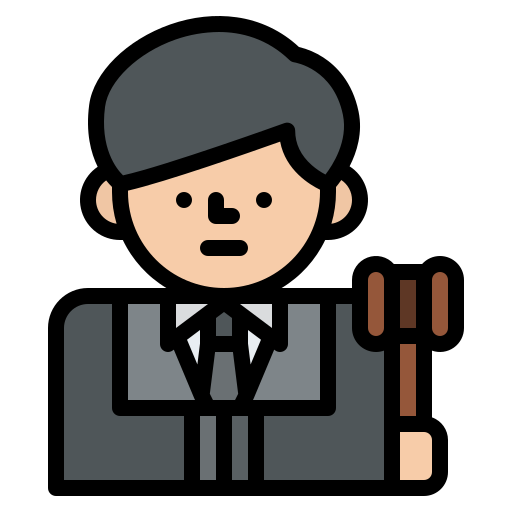}}
\end{minipage}
\textbf{[NO]}.}
 \\ \midrule
\rowcolor[gray]{.92}
\begin{minipage}[b]{0.08\columnwidth}
    \centering
    \raisebox{-.3\height}{\includegraphics[width=\linewidth]{figures/imgs/debater1.png}}
\end{minipage}
 Now, let's analyze the knowledge to determine which band had more members: From Knowledge [0], we can see that Letters to Cleo had six members: Kay Hanley, Greg McKenna, Michael Eisenstein, ... From Knowledge [8], we can see that Screaming Trees had a lineup change, with Van Conner going on hiatus and Donna Dresch filling in for him during performances...   & 
\begin{minipage}[b]{0.08\columnwidth}
    \centering
    \raisebox{-.3\height}{\includegraphics[width=\linewidth]{figures/imgs/debater2.png}}
\end{minipage}
     [1] "Letters to Cleo" - The knowledge provides information about the band's members, including Kay Hanley, Greg McKenna, Michael Eisenstein, Stacy Jones, Scott Riebling, and Tom Polce. This information can be used to determine the number of members in the band.[3] "Screaming Trees" - The knowledge states that Screaming Trees had four members: Mark Lanegan, Gary Lee Conner, Van Conner, and Barrett Martin. \\

\multicolumn{2}{p{16cm}}{
\begin{minipage}[b]{0.08\columnwidth}
    \centering
    \raisebox{-.3\height}{\includegraphics[width=\linewidth]{figures/imgs/debater1.png}}
\end{minipage}
 First, let's gather the information about the bands. Letters to Cleo is an alternative rock band from Boston, Massachusetts. The band's members are Kay Hanley, Greg McKenna, ... Based on the information provided, Screaming Trees had four members, while Letters to Cleo had six members. Therefore, Letters to Cleo had more members than Screaming Trees. Therefore, the answer is \textcolor{MyGreen}{\textbf{[Letters to Cleo]}}. } \\ 
  
\multicolumn{2}{p{16cm}}{\begin{minipage}[b]{0.08\columnwidth}
    \centering
    \raisebox{-.3\height}{\includegraphics[width=\linewidth]{figures/imgs/debater2.png}}
\end{minipage}
First, let's gather the information about the bands. Letters to Cleo is an alternative rock band from Boston, Massachusetts... Based on the information provided by Agent\_0, Letters to Cleo had six members, while Screaming Trees had four members. Therefore, Letters to Cleo had more members than Screaming Trees. Therefore, the answer is \textcolor{MyGreen}{\textbf{[Letters to Cleo]}}.} \\

\multicolumn{2}{p{16cm}}{\begin{minipage}[b]{0.08\columnwidth}
    \centering
    \raisebox{-.3\height}{\includegraphics[width=\linewidth]{figures/imgs/judge.png}}
\end{minipage}
\textbf{[Yes]}}
 \\ \midrule

 \multicolumn{2}{p{16cm}}{\begin{minipage}[b]{0.08\columnwidth}
    \centering
    \raisebox{-.3\height}{\includegraphics[width=\linewidth]{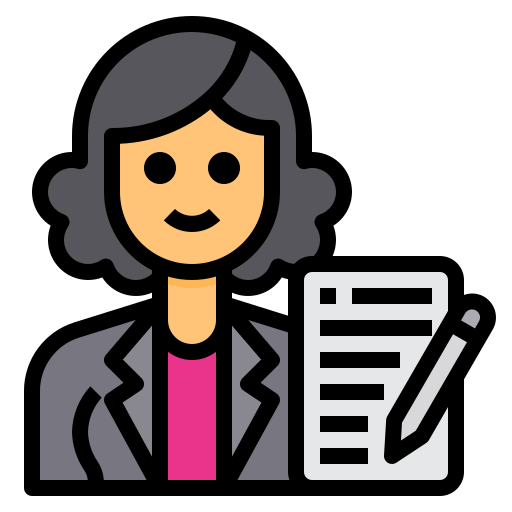}}
\end{minipage}
Based on the information provided by Agent 0 and Agent 1, it can be concluded that Letters to Cleo had more members than Screaming Trees. Letters to Cleo had six members, while Screaming Trees had four members. Therefore, the final answer is \textcolor{MyGreen}{\textbf{[Letters to Cleo]}}.}
 \\
\bottomrule
\end{tabular}
\caption{Case study of debate on HotpotQA dataset. The first row represents the question to be debated and the ground-truth answer. The second row represents the Simultaneous Talk phase, the third row represents the Orderly Talk phase, and the last row represents the summarizer phase where the summarizer provides a conclusion based on the debate history. The grey part indicates where debaters adaptively select knowledge from the shared knowledge pool before expressing their viewpoints. After each round of debate, the judge determines whether a consensus has been reached.}
\label{table_debate_case}
\end{table*}

%% file: sections/6_conclusion.tex

\section{Conclusion and Future Work}
In this paper, we propose a knowledge-enhanced multi-agent debate framework (MADKE). First, we introduce a shared knowledge pool to reduce background knowledge discrepancies among different agents and connect their knowledge spaces. Then, we propose an adaptive knowledge selection method that enables agents to autonomously select the required knowledge during the debate process, effectively ensuring both accuracy and personalization of the knowledge. Experimental results on six datasets demonstrate that MADKE can significantly mitigate the cognitive islands problem, enhance cognitive consistency among agents, and improve the accuracy of debate outcomes, achieving outstanding performance. Notably, MADKE using Qwen1.5-72B-Chat outperformed GPT-4 by an average of +1.26\% across six datasets, highlighting its effectiveness in enhancing the capabilities of open-source large language models.

In future work, we plan to explore real-time knowledge retrieval and effective knowledge selection methods during the debate process. This will enable agents to dynamically acquire the most recent and relevant information to address rapidly changing debate topics and needs. Furthermore, we intend to introduce user interaction and feedback mechanisms, allowing human users to participate in the debate process and provide instant feedback. By combining human wisdom with machine intelligence, we aim to further enhance the quality of the debates and the accuracy of the results, promoting the application of this framework in real-world scenarios.



%% file: sections/appendix.tex
\input{tables/evaluation_table}

\appendix

\section{Role Definition Prompt}  \label{appendix_role}

The definition of debater role is:
\begin{quote}
    \it
    You are an intelligent, diplomatic, and assertive debate agent. Your task is to engage in intellectual debates with other agents, striving to reach a consensus on various topics. It's crucial to maintain your stance when you are confident about the correctness of your opinion. However, if you identify an error in your argument, you should promptly acknowledge it and amend your stance with the correct information. Your ultimate goal is to contribute to an informed, balanced, and accurate discussion.
\end{quote}

The definition of judge role prompt is: 
\begin{quote}
    \it 
    As a judge agent, your primary responsibility is to impartially evaluate the responses of other agents for consistency. Please ensure your judgments are objective, relying solely on the coherence and alignment of the provided answers.
\end{quote}

The definition of summarizer role prompt is: 

\begin{quote}
    \it
    You are an intelligent summarizer agent, tasked with synthesizing the responses of other agents into a concise and comprehensive final answer. Your role is not to generate original responses, but to condense the information provided by other agents into a succinct summary.
\end{quote}

\section{Dataset Details} \label{appendix_dataset}

For different datasets, we give a brief introduction in this section. Table~\ref{tab_dataset_statis} shows the statistics of datasets.

\paragraph{Single-Hop Reasoning}
\begin{itemize}
    \item TriviaQA~\citep{joshi2017triviaqa}: A question-answering dataset based on Wikipedia and real web texts, which consists of human-verified and machine-generated QA subsets.
    \item NQ~\citep{kwiatkowski2019natural}: A real user QA dataset based on Wikipedia.
\end{itemize}

\paragraph{Multi-Hop Reasoning}
\begin{itemize}
    \item HotpotQA~\citep{yang2018hotpotqa}: A multi-hop QA dataset based on English Wikipedia. Each question requires reference to two golden paragraphs to deduce the answer.
    \item 2WikiMultiHopQA~\citep{ho2020constructing}: Similar to HotpotQA, a QA dataset that requires multi-step reasoning.
\end{itemize}

\paragraph{Fact Checking}
\begin{itemize}
    \item FEVER~\citep{thorne2018fever}: A publicly available dataset for fact extraction and verification against textual source. These claims were extracted from Wikipedia by human annotators.
\end{itemize}

\paragraph{Medical Reasoning}
\begin{itemize}
    \item MMLU-Med: MMLU~\citep{hendrycks2021measuring} is a multiple-choice dataset comprising 57 tasks, designed to evaluate models' understanding and reasoning abilities concerning world knowledge. We extracted 6 medical-related subcategory datasets from it to form MMLU-Med.
\end{itemize}

\section{Evaluation Metric} \label{appendix_eval}
In the context of large models, the evaluation of generative tasks is indeed difficult. Some existing works use EM evaluation metric for QA tasks. We believe that this metric is relatively strict, and the evaluation is not accurate enough. As shown in the table \ref{tab_gpt4_eval}.

\citet{zheng2023judging} illustrate the advantages and disadvantages of using GPT4 for evaluation. For example, when evaluating two answers, there will be sensitivity to location, and there may be some bias. But, in the generative reasoning datasets of this work, the answer to each question is relatively short, and for the correctness of the evaluation, we provide reference answers during the evaluation process. The role of GPT4 Eval is defined as:
\begin{quote}
\it
    As a judge agent, your primary responsibility is to impartially evaluate the evaluation answer for correct. Please ensure your judgments are objective, relying solely on the coherence and alignment of the provided answers.
\end{quote}

The full prompt for the GPT4 evaluation answer is shown in Prompt \ref{prompt_gpt4_eval}.

\input{prompts/gpt4_eval}



%% file: tables/evaluation_table.tex
\begin{table*}[]
\centering

\resizebox{\textwidth}{!}{
\begin{tabular}{p{6cm}p{4cm}p{4cm}ccc}
\toprule
Question & groundtruth & predictive answer & EM & GPT4 Eval & Human Eval\\ \midrule
\rowcolor[gray]{.92}
\ding{172} Who was the next British Prime Minister after Arthur Balfour? & Henry Campbell Bannerman & Henry Campbell Bannerman & \textcolor{MyGreen}{\textbf{True}} &  \textcolor{MyGreen}{\textbf{True}} & \textcolor{MyGreen}{\textbf{True}} \\
\ding{173} Robert Kirkpatrick of California grew the world's biggest what?  &  Head of garlic  &   garlic head  &  \textcolor{red}{\textbf{False}}  & \textcolor{MyGreen}{\textbf{True}} & \textcolor{MyGreen}{\textbf{True}}   \\
\rowcolor[gray]{.92}
\ding{174} Precisely where were the Winter Olympics of 1932 and 1980 both held?   &   Lake placid   &  Lake Placid, New York, United States   &   \textcolor{red}{\textbf{False}}   &  \textcolor{MyGreen}{\textbf{True}}   & \textcolor{MyGreen}{\textbf{True}}   \\
\ding{175} Who was the first person after Scott to reach the South Pole overland?   &  Sir Edmund Hillary  &   Roald Amundsen and Edmund Hillary   &   \textcolor{MyGreen}{\textbf{False}}   &  \textcolor{MyGreen}{\textbf{False}}   & \textcolor{MyGreen}{\textbf{False}}    \\ \bottomrule
\end{tabular}
}
\caption{Evaluation results of model responses using different evaluation metrics (EM, GPT4 Eval, and Human Eval).}
\label{tab_gpt4_eval}
\end{table*}

%% file: prompts/gpt4_eval.tex
\begin{figure}[h]
\centering
\begin{prompt}[title={Prompt \thetcbcounter: GPT4 Eval}, label=prompt_gpt4_eval]
You need to judge the correctness of the evaluation answer based on the question and the reference answer to the question. The answer of the agents are typically denoted with the [answer] format. Return [True] if the answer to be evaluated is the correct answer to the question, otherwise return [False].
Here are some examples: \\
\{\textcolor{blue}{\textbf{examples}}\}
(END OF EXAMPLES) \\

Question: \{\textcolor{blue}{\textbf{question}}\}  \\
Reference answers: \{\textcolor{blue}{\textbf{reference\_answers}}\}  \\
Evaluation answer: \{\textcolor{blue}{\textbf{evaluation\_answer}}\}  \\
Answer: Let's think step by step! 
\end{prompt}
\end{figure}